\documentclass{article}
\usepackage{ijcai21}

\usepackage{times}
\usepackage{soul}
\usepackage{url}
\usepackage[hidelinks]{hyperref}
\usepackage[utf8]{inputenc}
\usepackage[small]{caption}
\usepackage{graphicx}
\usepackage{amsmath}
\usepackage{amsthm}
\usepackage{booktabs}
\usepackage{algorithm}
\usepackage{algorithmic}
\usepackage{color,xcolor}
\usepackage{multirow}
\usepackage{mfirstuc}
\makeatletter
  
\def\ie{\emph{i.e., }}

\def\targetproto{class-level prototype}

\def\generalproto{prototype} 
\def\prototype{\generalproto{}}

\def\generatedproto{episodic prototype} 

\def\Targetproto{Class-level prototype}

\makeatother

\title{Learning Class-level Prototypes for Few-shot Learning}
\author{
    Anonymous IJCAI 2021 submission
    \affiliations
    Paper ID: 1316
    \emails
}

\iftrue
\author{
Minglei Yuan$^1$
\and
Wenhai Wang$^1$\and
Tao Wang$^1$\and
Chunhao Cai$^1$\and
Qian Xu$^1$\And
Tong Lu$^{*1}$
\affiliations
$^1$Nanjing University\\
\emails 
mlyuan@smail.nju.edu.cn}
\fi

\begin{document}

\maketitle

\begin{abstract}
Few-shot learning aims to recognize new categories using very few labeled samples. 
Although few-shot learning has witnessed promising development in recent years, most existing methods adopt an average operation to calculate prototypes, thus limited by the outlier samples.
In this work, we propose a simple yet effective framework for few-shot classification, which can learn to generate preferable prototypes from few support data, with the help of an episodic prototype generator module. The generated prototype is meant to be close to a certain \textit{\targetproto{}} and is less influenced by outlier samples. 
Extensive experiments demonstrate the effectiveness of this module,
and our approach gets a significant raise over baseline models, and get a competitive result compared to previous methods on \textit{mini}ImageNet, \textit{tiered}ImageNet, and cross-domain (\textit{mini}ImageNet $\rightarrow$ CUB-200-2011) datasets.

\end{abstract}

\section{Introduction}
\label{sec:introduction}

Although few-shot learning has witnessed promising development in recent years, most existing methods adopt an average operation to calculate prototypes, thus limited by the outlier samples.
In this work, we propose a simple yet effective framework for few-shot classification, which can learn to generate preferable prototypes from few supporting data, with the help of an episodic prototype generator module. The generated prototype is meant to be close to a certain \textit{\targetproto{}} and is less influenced by outlier samples. 
Extensive experiments demonstrate the effectiveness of this module,
and our approach gets a significant raise over baseline models, and get a competitive result compared to previous methods on \textit{mini}ImageNet, \textit{tiered}ImageNet, and cross-domain (\textit{mini}ImageNet $\rightarrow$ CUB-200-2011) datasets.

The goal of few-shot learning is to learn new categories from limited labeled data. In past years, many methods~\cite{matching,protonet,maml,rn,denseclassification,leo,pacbayesian} have been proposed to solve the few-shot classification problem.
Among them, some effective solutions~\cite{rn,protonet,subspace,matching} are to learn the embedding of support sets and query sets and then use the embedding distance to classify the query samples.
For example, ProtoNet (Prototypical Networks)~\cite{protonet} makes a few-shot classification by measuring the Euclidean distance between the query embedding and the mean embedding of 
each class in support set. 
RelationNet~\cite{rn} develops a learnable module that learns to compare the query image against few-shot labeled sample images.

\begin{figure}[tb] 
\centering 
\includegraphics[width=0.5\textwidth]{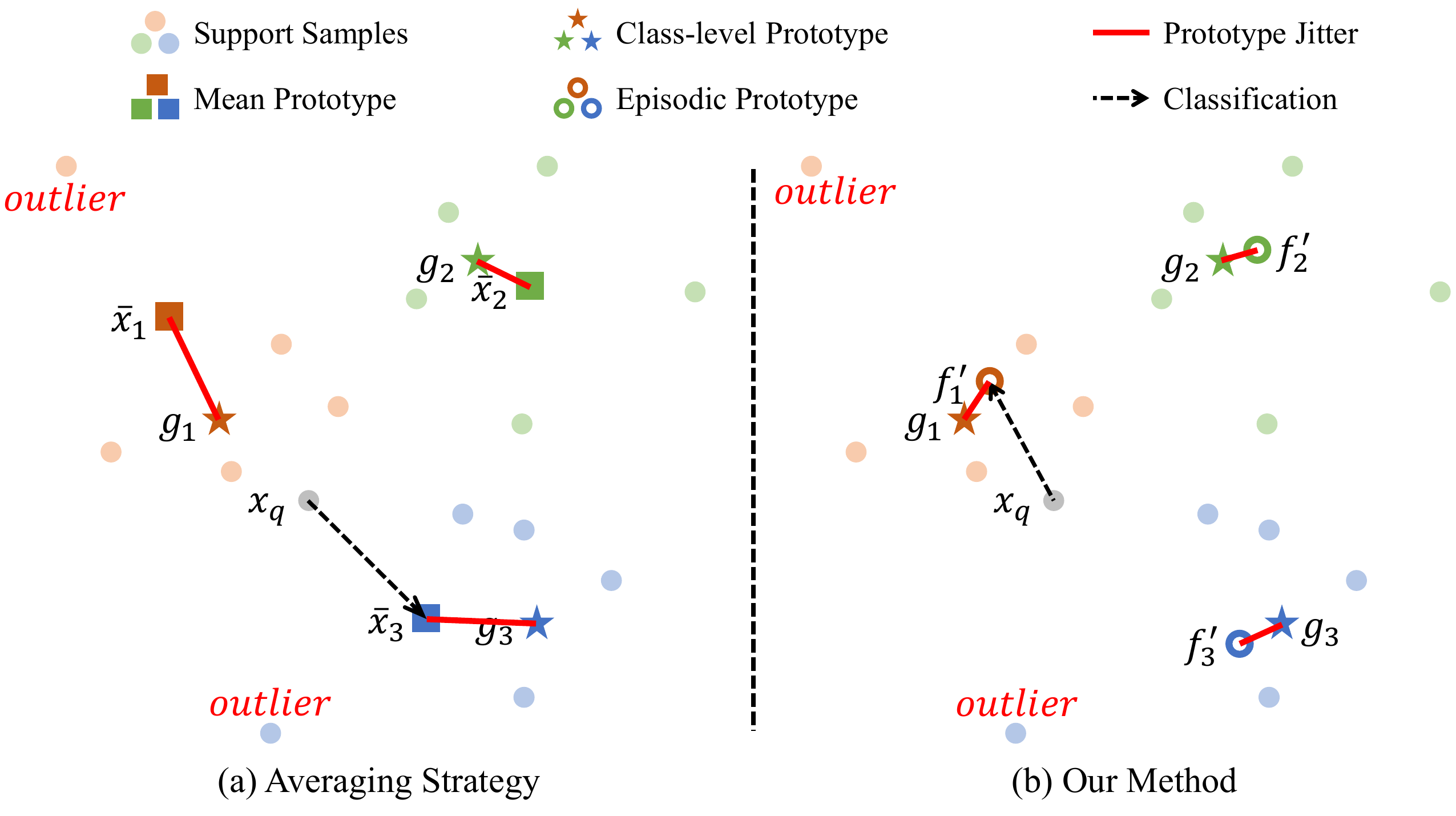} 
\caption{The class-level prototype jitter phenomenon. Class-level prototypes in a 3-way 5-shot task are subject to jitter due to outliers.
The classification of query samples $x_q$ is achieved by measuring the distance between different classes of prototypes. 
(a) $x_q$ is misclassified due to the influence of outlier points, as the averaging strategy is easily affected by the existence of outliers.
(b) Our method generates episodic prototypes that are closer to the class-level prototypes, fixing classification errors.} 
\label{fig_motivation}
\end{figure}

Although many few-shot classification algorithms have achieved promising performance, they rarely concentrate on designing a good strategy to generate class-level prototypes to represent each class in the support set.
Some recent methods~\cite{rn,protonet,tadam,denseclassification,lowshortimaginarydata} adopt the averaging strategy (\ie using the arithmetic mean of embeddings of support samples in each class) to approximate class-level prototypes.
Although this strategy works well in ideal situations where sufficient samples are given and each of them is independently sampled from the same distribution~\cite{khintchinelay}.
However, in a few-shot classification, the influence of one single outlier on the averaging strategy may be significant because there are few supporting samples. 
This phenomenon is named as \textit{class-level prototype jitter} in ~\cite{ifsm}, which is shown in Figure~\ref{fig_motivation}.
 
%

 
 To remedy this problem, several methods~\cite{subspace,projectivesubspace} choose a preferable class-level prototype from the linear subspace constructed by the support set, whose main spirit is to calculate a subspace of the feature space for each category, and then project the feature vector of the query sample into the subspace, and take the distance measurement in the subspace as the criterion for predicting the category. An iterative process is proposed in IFSM~\cite{ifsm} to optimize prototypes, which leads to high time complexity due to complex iterations. In addition, PSN~\cite{projectivesubspace} and IFSM~\cite{ifsm} do not work on 1-shot tasks.
 And all previous methods omit the guidance of class-level prototypes, which is the average feature of a large number of samples of the same class in the training set.
 
 To solve this problem, we propose an \textit{episodic prototype generator} to extract the prototype supervised with the \textit{class-level prototype}.
 In this way, the proposed episodic prototype generator needs to generate a prototype that can converge to a more robust prototype. This reduces the impact of outliers in the prototype calculation.

Specifically, we propose an effective framework for the few-shot classification task.
The proposed method consists of three main modules: the \targetproto{} generator module, the episodic prototype generator module, and the metrics module.
Among them, the \targetproto{} generator module aims to learn class-level prototypes (\ie the global mean of sample features for each class in the training set) from the training data so that our algorithm can directly benefit from the class-level prototype when dealing with few-shot classification problem.
The episodic prototype generator is a learnable module constructed with a Transformer \cite{transformer}. It can help the network capture the contextual information from samples with the same category.
Besides, the output of the episodic prototype generator is supervised with \targetproto{}s produced by \targetproto{} generator module. In this way, the proposed episodic prototype generator model can tender to get the prototype closer to the \targetproto{}s and reduce the impact of outlier points. 
Finally, the \textit{metric module} determines the category to which the query sample belongs based on the distance. 

Extensive experiments demonstrate that the proposed method is competitive with state-of-the-art few-shot learning methods.
Firstly, we directly use \targetproto{} for classification and get a surprising result with an accuracy of 89.15\%. The experiment proves it is wise to choose it as the classification basis. Secondly, we conduct consistent comparative experiments on two datasets to compare several representative methods of few-shot classification. We achieve the state-of-the-art performance on \textit{tiered}ImageNet 5-way 1-shot and 5-shot, and on \textit{mini}ImageNet 5-way 1-shot, with accuracy of 71.78\%, 85.91\% and 66.68\%, respectively. Finally, we further evaluate the performance of this method on cross-domain tasks. The fact shows that our model has a strong ability to migrate and is highly competitive with other methods.

The contributions of our work are as follow:
\begin{enumerate}
    \item We assume that the \targetproto{}s can be good candidates for class-level prototypes in few-shot classification, and justify it through ablation experiments.
    \item We propose a mechanism which takes support samples as input and generates class-level prototypes relatively close to corresponding \targetproto{}s. 
    \item We show that the proposed mechanism brings a significant rise to the performance of various few-shot learning methods on \textit{mini}ImageNet, \textit{tiered}ImageNet and a cross-domain task (\textit{mini}ImageNet $\rightarrow$ CUB-200-2011).
    
\end{enumerate}

\section{Related Works}
\label{sec:related_works}
Few-shot learning has been of great interest for a long time, and a large quantify of few-shot learning algorithms have been proposed in the literature. These methods can be approximately divided into two categories. The first kind of approach aims at strategies of constructing a general classifier. They show feasible performance on few-shot tasks. Concretely, this can be done by inventing a certain data augmentation algorithm~\cite{ctm} or optimization strategy based on the content of the support and query sets~\cite{matching}. This can also be achieved by directly predicting the weights of the classifier~\cite{leo}.

The other school of few-shot learning methods~\cite{protonet,matching,rn,ctm,ifsm,subspace} focuses on learning a feature extractor to embed all samples in the support set and the query set, followed by a classifier making its decisions based on the embedding of query samples and embeddings of support samples. These methods are usually called metric-based few-shot learning methods. Early methods of this category directly compare the embedding of query samples to embeddings of all support samples, which are similar to weighted nearest-neighbor classifiers like Matching Network~\cite{matching}. Prototypical Network~\cite{protonet} decides that it is more efficient to use \textit{prototypes}, \ie embeddings of classes to compare with embeddings of query samples. 

Besides, merely using the arithmetic mean as prototypes are far from feasible. Thus, several methods have been proposed to alleviate this problem through different means. For instance, CTM~\cite{ctm} introduces a category traversal module for generating \prototype{}s, which can explicitly extract intra-class commonality and inter-class uniqueness. IFSM~\cite{ifsm} assumes a relatively feasible prototype can be represented by a linear combination of embeddings of support samples from that class and adopts an iterative process to gradually optimize that linear combination. Subspace Network~\cite{subspace} is based on a similar assumption. However, it generates an independent set of prototypes for each query sample and is limited to certain types of distance metrics. 

Our work is somehow based on a similar idea compared to these methods, which is to generate \prototype{}s through a learnable neural network. However, our work is different in that our model is supervised by the \targetproto{}, which is a good candidate for the \prototype{}. This facilitates us to obtain a \prototype{} that is closer to \targetproto{}.

\begin{figure*}[t] 
\centering 
\includegraphics[width=0.85\textwidth]{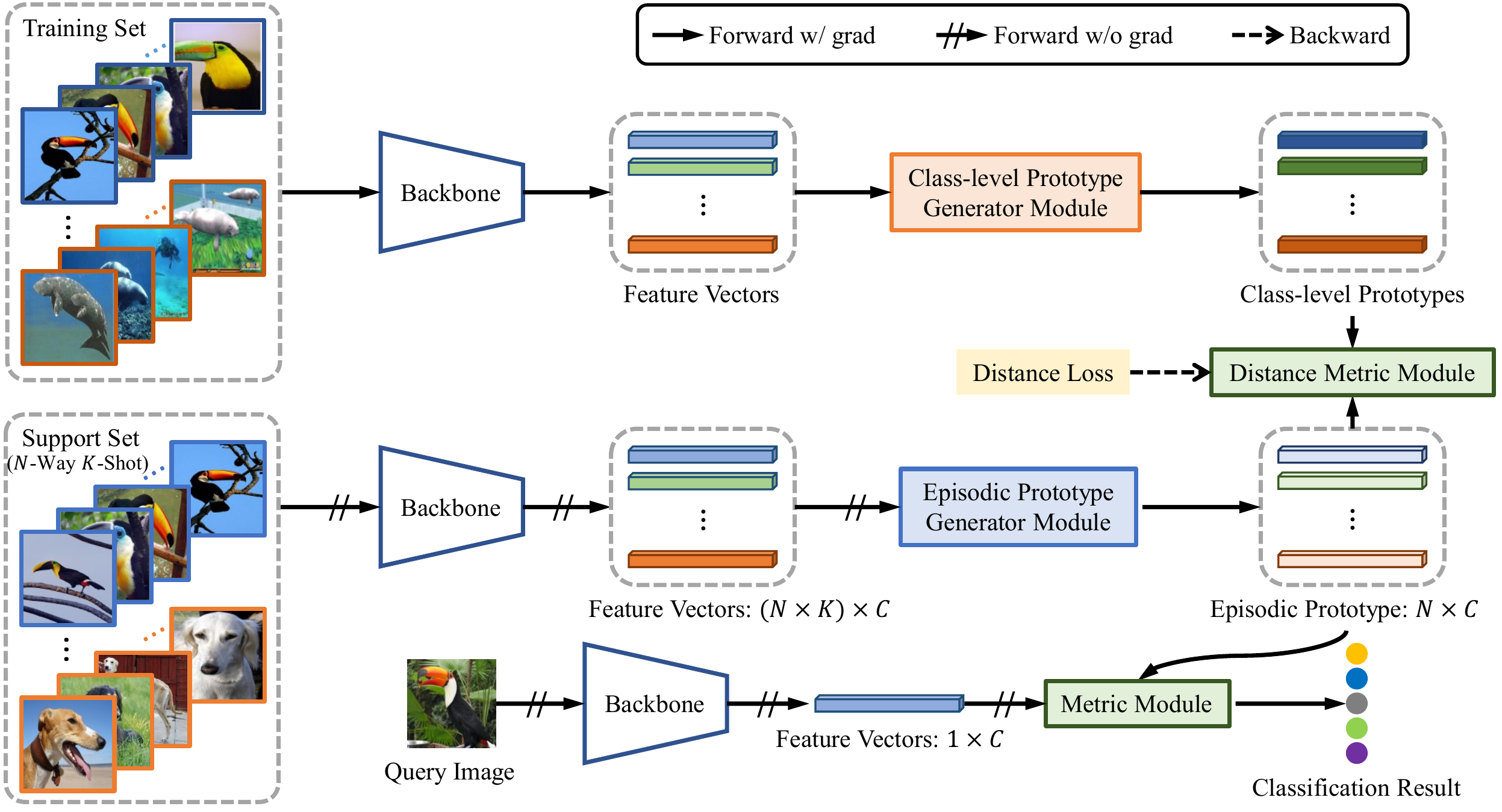} 
\caption{The overview of our proposed framework for few-shot learning, which is mainly composed of a \targetproto{} generator module, an episodic prototype generator module, and a metric module. } 
\label{fig_framework}
\end{figure*}

\section{Preliminary}
In this section, we introduce some notation and formalization of few-shot learning in an image classification task. Let $(x, y)$ denotes a labeled example where $x$ is a sample and $y$ is the corresponding ground-truth. Datasets are often divided into a base class dataset $\mathcal{D}_{train}$ and a novel class dataset $\mathcal{D}_{novel}$, and $\mathcal{D}_{train}$ and $\mathcal{D}_{novel}$ do not intersect. The purpose of few-shot learning is to build a good meta-learner $\Phi_{\theta}$ for the novel classes in $\mathcal{D}_{novel}$ with a few labeled samples (or called supported samples).

Meta-learning \cite{matching} is an effective way for few-shot learning, which can utilize the training set in an episodic manner by mimicking the few-shot learning setting. Meta-learning consists of two phases: meta-training and meta-test. 

In the meta-training phase, data is randomly sampled from $\mathcal{D}_{train}$ to simulate a scenario $\mathcal{T}$, and the few-shot learning meta-learner $\Phi_{\theta}$ is trained with thousands of randomly sampled scenarios.
In each scenario, the sampled data are divided into a support set $\mathcal{S}$ and a query set $\mathcal{Q}$, which satisfies that the elements of $\mathcal{S}$ and $\mathcal{Q}$ do not intersect, but are drawn from the same class. The support set $\mathcal{S}$ and the query set $\mathcal{Q}$ are formalized as follow: 
\begin{equation}
	\label{eq_support_set}
	\begin{split}
	\mathcal{S} = \{x_i | y_i = k,(x_i,y_i) \in \mathcal{D}_s \}, \\
	\mathcal{Q} =  \{x_i | y_i = k,(x_i,y_i) \in \mathcal{D}_q \}.
    \end{split}
\end{equation}
In the meta-test phase, the learned few-shot learning meta-learner $\Phi_{\theta}$ is tested with a large number of episodes drawn from $\mathcal{D}_{novel}$, and the average prediction accuracy is used to measure the learning ability of $\Phi_{\theta}$.


In this paper, we consider only the $N$-way $K$-shot few-shot learning paradigm, where each support set $\mathcal{S}$ contains $N$ classes, each class includes $K$ samples, and the query set $\mathcal{Q}$ consists of unlabeled samples belonging to the classes of support set $\mathcal{S}$. 

Meta-learner $\Phi_{\theta}$ is trained to fit few-shot classification tasks by minimizing the prediction loss on the query set $\mathcal{Q}$ as in Eq. (\ref{eq_optim}).
\begin{equation}
    \label{eq_optim}
    \underset{\theta}{\operatorname{arg~min~}} \frac{1}{N_q} \sum_{(x_i, y_i) \in \mathcal{Q}}\mathcal{L}(y_i, \Phi_{\theta}(x_i, \mathcal{S})) ,
\end{equation}
where $N_q$ is the number of samples in $\mathcal{Q}$, and $\mathcal{L}$ is the loss function. The meta-learner $\Phi_{\theta}$ is required to train with thousands of randomly sampled tasks under the constraint of the loss function $\mathcal{L}$.

However, the proposed framework is trained using distance loss. The details are given in Section \ref{sec:method}.

\section{Method}
\label{sec:method}
\subsection{Framework}
The key idea of our model is to elaborate how to get a better \generalproto{} supervised with a \targetproto{} in case of insufficient support samples.
To overcome the class-level prototype jitter problem, we propose a pipeline to fit \targetproto{} using few support samples. Figure \ref{fig_framework} shows the overall architecture of our proposed model. The proposed algorithm consists of three main components: a \textit{\targetproto{} generator module}, an \textit{episodic prototype generator module} and a \textit{metric module}. 

As illustrated in Figure \ref{fig_framework}, \targetproto{} works in three steps. The training data are first fed into the \targetproto{} generator module.
A generic classifier is first trained with the training set and the parameters of the classifier are frozen. Then, the features of all samples in each category are obtained using the trained classifier.
Finally, the arithmetic means of all sample features in each category are used as \targetproto{}. 
\Targetproto{} generator module provides a backbone with fixed parameters for the following episodic prototype generator module. 
The episodic prototype generator is designed to produce prototype that is closer to \targetproto{}, which is detailed in Section \ref{sec_prototy_feature_geanerator}. The metric module accomplishes the recognition of images by measuring the distance between the \generatedproto{} features and the feature vectors of images in query set. We will introduce the details of these modules in the following.

\subsection{Class-level Prototype Generator Module}
The \targetproto{} generator module aims at generating the average of the features of all samples in the same class. 
Existing few-shot learning methods are usually trained and tested on \textit{mini}ImageNet~\cite{matching} and \textit{tiered}ImageNet~\cite{tieredimagenet}. Each class of the training set $\mathcal{D}_{train}$ consists of a considerable number of samples. In this paper, we consider the global class prototype as the best candidate for the class-level prototype. And then propose the \targetproto{} to optimize the \generatedproto{}. Specific optimization method will be described in detail in Section \ref{sec_prototy_feature_geanerator}. We focus on how to calculate \targetproto{} in this part.

To obtain the \targetproto{}, a classifier is first trained using $\mathcal{D}_{train}$. The classifier consists of a backbone network $\mathcal{F}_\theta$ and a fully connected network with a softmax layer and is optimized using cross-entropy loss.
Specifically, the backbone network $\mathcal{F}_\theta$ is used to extract the features of images with a type of residual network \cite{resnet}. The detailed structure of the backbone network $\mathcal{F}_\theta$ is presented in Section \ref{sec:impementation_detail}.

Then, the trained backbone network $\mathcal{F}_{\theta}$ is used to extract the features of samples in each class in the training set, and the arithmetic mean of all features in the same class is adopted as \targetproto{}.
\begin{equation}
    g_i =  \frac{1}{M} \sum_{j=1}^{M} \mathcal{F}_{\theta}(x_{ij}),
\end{equation}
where $M$ is the total number of samples in the \emph{i}-th class, $x_{ij}$ indicates the input sample.

\subsection{Episodic Prototype Generator Module}
\label{sec_prototy_feature_geanerator}
When calculating the prototype of an episode using the arithmetic mean in a few-shot learning scenario, the prototypes are often affected by outlier points. In this section, we propose an episodic prototype generator module, which is designed to narrow the difference between the \generalproto{} and the \targetproto{}.  The proposed episodic prototype generator generates a \generalproto{} under the supervision of a \targetproto{}, and the final optimization goal is to obtain a \generalproto{} that is closer to the \targetproto{}.

To achieve this, in each episode, the samples $\{x_{i1}, \cdots, x_{iK}\}$ in the same class are first fed into the pre-trained feature extractor $\mathcal{F}_\theta$, which can obtain their feature vectors, denoted as $f_{ij} = \mathcal{F}_\theta(x_{ij})$. Then, the extracted feature vectors $\{f_{i1}, \cdots, f_{iK}\}$ are fed into the episodic prototype generator $\mathcal{G}_{\gamma}$ and produce \generatedproto{} $f'_i$, denoted as Eq \ref{eq_feature_generate}.
\begin{equation}
    f'_{i} = \mathcal{G}_{\gamma}(f_{i1}, \cdots, f_{iK}),
    \label{eq_feature_generate}
\end{equation}
where $\gamma$ is the parameters of the episodic prototype generator $\mathcal{G}$, $i$ indicates the \emph{i}-th category.

Finally, we compute the distance between the \generatedproto{} $f'_i$ and \targetproto{} $\hat{f}_i$ and use the distance loss to optimize the parameters $\gamma$ of the episodic prototype generator $\mathcal{G}_{\gamma}$. Euclidean distance is adopted to measure the distance loss in this paper.

\begin{equation}
    \mathcal{L}_{distance} = \frac{1}{N} \sum_{i=1}^{N} \mathcal{M}(f'_{i}, \hat{f}_{i}),
\end{equation}
where $\mathcal{M}$ is a distance metric module, $N$ indicates the number of categories in an episode, $\hat{f}_{i}$ is the \targetproto{} of the \emph{i}-th category in an episode.

In this paper, we employ a multi-head self-attention mechanism~\cite{transformer} as the feature generator $\mathcal{G}$, which considers the contextual information of samples in the same episode. The multi-head self-attention mechanism outputs the refined feature vectors of the input samples. We use the mean of updated feature vectors as prototypes. These prototypes can converge to the \targetproto{} with the help of the loss function $\mathcal{L}_{distance}$, thus eliminating the effect caused by outliers in input samples. 


\subsection{Metric Module}
The metric module $\mathcal{M^{'}}$ classifies the images in the query set. First, the trained backbone network $\mathcal{F}_\theta$ is used to extract the features of samples in the query set. Then, the metric module $\mathcal{M^{'}}$ classifies the query samples according to the distance between the features of images in the query set and the \generatedproto{} features $\{f'_1, \cdots, f'_N\}$. In this paper, Euclidean distance is chosen as the metric to obtain the classification accuracy. 

\begin{table*}[ht]
    \centering
     \caption{\textbf{5-way few-shot classification accuracy (\%) on \textit{mini}ImageNet and \textit{tiered}ImageNet.} We report the average accuracy and 95\% confidence interval of 600 randomly generated test sets for each task. The superscript $^{\dag}$ indicates that the experimental results are from \protect\cite{closerfewshot}. \textcolor{red}{Red}: the best; \textcolor{blue}{Blue}: the second best.}
    \scalebox{0.75}{
    \begin{tabular}{r|c|c|c c|c c}
    \toprule
        \multirow{2}*{\textbf{Methods}}
         &\multirow{2}*{\textbf{Backbone}}
         &\multirow{2}*{\textbf{Source}}
         &\multicolumn{2}{c}{\textbf{\textit{mini}ImageNet}} 
         &\multicolumn{2}{c}{\textbf{\textit{tiered}ImageNet}} \\
         \multicolumn{1}{c|}{~}
         &\multicolumn{1}{c|}{~}
         &\multicolumn{1}{c|}{~}
         &\multicolumn{1}{c}{5-way 1-shot}
         &\multicolumn{1}{c}{5-way 5-shot}
         &\multicolumn{1}{c}{5-way 1-shot}
         &\multicolumn{1}{c}{5-way 5-shot} \\
    \midrule
    MatchNet \cite{matching} & 4CONV & NeurIPS16 & 46.6 & 60.0 & - & -\\
    ProtoNet \cite{protonet} & 4CONV & NeurIPS17 & 49.42 $\pm$ 0.78 & 68.20 $\pm$ 0.66 & 53.31 $\pm$ 0.89 & 72.69 $\pm$ 0.74\\
    MAML \cite{maml} & 4CONV & ICML17 & 48.70 $\pm$ 1.84 & 63.11 $\pm$ 0.92 & - & -\\
    RelationNet \cite{rn} & 4CONV & CVPR18 & 50.44 $\pm$ 0.82 & 65.32 $\pm$ 0.70 & 54.48 $\pm$ 0.93 & 71.32 $\pm$ 0.78 \\
    PSN~\cite{projectivesubspace} & 4CONV & ICLR19 & - & 66.62 $\pm$ 0.69 & - & -\\ 
    PAC-Bayesian \cite{pacbayesian} & 4CONV & TPAMI20 & 52.11 $\pm$ 0.43 & 63.87 $\pm$ 0.35 & - & -\\
    PN+IFSM~\cite{ifsm} & 4CONV & ICPR20 & - & 66.98$ \pm$ 0.68 & - & 78.81 $\pm$ 0.66 \\
    
    DC \cite{denseclassification}  & ResNet-12 & CVPR19 & 61.26 $\pm$ 0.20 & 79.01 $\pm$ 0.13 & - & - \\
    ProtoNet+TRAML \cite{prototraml} & ResNet-12 & CVPR20 & 60.31 $\pm$ 0.48 & 77.94 $\pm$ 0.57 & - & -\\
    MetaOptNet \cite{metaoptnet} & ResNet-12 & CVPR19 & 64.09 $\pm$ 0.62 & 80.00 $\pm$ 0.45 & 65.81 $\pm$ 0.74 & 81.75 $\pm$ 0.53 \\
    Ravichandran et al. \cite{ravichandranetal} & ResNet-12 & ICCV19 & 60.71 & 77.26 & 66.87 & 82.43\\
    Y. Tian et al. \cite{y.tianetal} & ResNet-12 & ECCV20 & 62.02 $\pm$ 0.63 & 79.64 $\pm$ 0.44 &69.74 $\pm$ 0.72 & \textcolor{blue}{84.41 $\pm$ 0.55}\\
    DSN-MR~\cite{subspace} & ResNet-12 & CVPR20 & 64.60 $\pm$ 0.72 & 79.51 $\pm$ 0.50 & 67.39 $\pm$ 0.82 & 82.85 $\pm$ 0.56\\

    Baseline \cite{closerfewshot} & ResNet-18 & ICLR19 & 51.75 $\pm$ 0.80 & 74.27 $\pm$ 0.63 & - & -\\
    Baseline++ \cite{closerfewshot} & ResNet-18 & ICLR19 & 51.87 $\pm$ 0.77 & 75.68 $\pm$ 0.63& - & - \\
    MatchNet\dag \cite{matching} & ResNet-18 & ICLR19 & 52.91 $\pm$ 0.88 & 68.88 $\pm$ 0.69 & - & -\\
    ProtoNet\dag \cite{protonet} & ResNet-18 & ICLR19 & 54.16 $\pm$ 0.82 & 73.68 $\pm$ 0.65 & - & - \\
    MAML\dag \cite{maml} & ResNet-18 & ICLR19 & 49.61 $\pm$ 0.92 & 65.72 $\pm$ 0.77 & - & -\\
    RelationNet\dag \cite{rn} & ResNet-18 & ICLR19 & 52.48 $\pm$ 0.86 & 69.83 $\pm$ 0.68 & - & -\\
    IDeMe-Net \cite{ideme} & ResNet-18 & CVPR19 & 59.14 $\pm$ 0.86 & 74.63 $\pm$ 0.74 & - & -\\
    Robust 20-dist++ \cite{robust20dist++} & ResNet-18 & ICCV19 & 59.48 $\pm$ 0.62 & 75.62 $\pm$ 0.48 & - & -\\
    
    Su et al. \cite{suetal} & ResNet-18 & ECCV20 & - & 76.6 & - & -\\
    AFHN \cite{afhn} & ResNet-18 & CVPR20 & 62.38 $\pm$ 0.72 & 78.16 $\pm$ 0.56 & - & -\\
    MTL+E3BM \cite{mtle3bm} & ResNet-25 & ECCV20 & 64.3 & 81.0 & 70.0 & 85.0 \\
    CC+rot \cite{cc+rot} & WRN-28-10 & ICCV19 & 62.93 $\pm$ 0.45 & 79.87 $\pm$ 0.33 & 62.93 $\pm$ 0.45 & 79.87 $\pm$ 0.33\\
    LEO \cite{leo} & WRN-28-10 & ICLR19 & 61.76 $\pm$ 0.08 & 77.59 $\pm$ 0.12 & 66.33 $\pm$ 0.05 & 81.44 $\pm$ 0.09\\
    Centroid alignment \cite{alignment} & WRN-28-10 & ECCV20 & \textcolor{blue}{65.92 $\pm$ 0.60} & \color{red}82.85 $\pm$ 0.55 & - & -\\
    
    PAC-Bayesian \cite{pacbayesian} & no-standard & TPAMI20 & 63.73 $\pm$ 0.57 & 78.04 $\pm$ 0.45 & \textcolor{blue}{70.82 $\pm$ 0.30} & 81.84 $\pm$ 0.21\\
    \midrule
    Ours & ResNet-12 & - & \textcolor{red}{66.68$\pm$0.81} & \textcolor{blue}{82.22 $\pm$ 0.56} & \textcolor{red}{71.78 $\pm$ 0.91} & \textcolor{red}{85.91$\pm$0.60} \\

    \bottomrule
    \end{tabular}
    }
    \label{tab:miniacc}
\end{table*}

\section{Experiments}

\subsection{Datasets}
\label{sec:datasets}
\paragraph{\textit{mini}ImageNet.}It contains 100 classes with 600 images in each class. Following the previous work~\cite{rn,protonet}, we divide this dataset into training, validation, and testing sets, with 64, 16, and 20 classes, respectively.

\paragraph{\textit{tiered}ImageNet.} It consists of a total of 779,165 images with 608 classes. Each image in \textit{tiered}ImageNet has a size of $84\times84$. It is a much more challenging dataset because of the realistic regime of test classes that are less similar to training ones. In our experiments, we follow the setting of ~\cite{ctm} where $351$ classes are used to make up the training set, 97 for validation and 160 for testing.

\paragraph{\textit{mini}ImageNet $\rightarrow$ CUB-200-2011.}To evaluate the effects of the proposed algorithm in a cross-domain scenario, we use the \textit{mini}ImageNet  as our training set and the validation set and testing set are from CUB-200-2011 \cite{cub}. The \textit{mini}ImageNet involves generic objects, while CUB-200-2011 contains 200 classes and 11,788 fine-grained bird images. This dataset mainly validates the cross-domain learning capability of the model.


\subsection{Implementation Details}
\label{sec:impementation_detail}

\paragraph{Backbone.}We adopt a 12-layer residual network (ResNet-12)~\cite{resnet} as the backbone network of our algorithm~\cite{denseclassification,metaoptnet,dsnmr}. ResNet-12 is composed of four residual blocks~\cite{resnet}.
Each residual block contains three $3\times3$ convolutional layers and a shortcut connection. At the same time, every convolutional layer is followed by a BathNorm layer and a LeakyReLU activation function (negative slope 0.1). In our experiments, the backbone network is pre-trained on the training set  $\mathcal{D}_{base}$ like \cite{leo}. 


\paragraph{Training Schema.} As many few-shot learning methods~\cite{protonet,rn}, we utilize a meta-training strategy to train our model. Specifically, in each N-way K-shot episode, we randomly sample K images per class to make up a support set, and $15$ images per class for the query set.  The total training process lasts $200$ epochs, and each epoch contains 200 episodes. We adopt stochastic gradient descent (SGD) as the optimizer. The learning rate is initially set to $0.01$ and is multiplied by $0.618$ whenever the validation score does not improve in $7$ consecutive epochs.

\paragraph{Evaluation protocols.}We evaluate our model on the validation at the end of each epoch and verify the performance of our model in the testing set. Specifically, we test the models with 600 randomly sampled tasks and calculate the mean and 95\% confidence interval of the prediction accuracy. In each sampling task, we randomly sampled 15 images for each category.

\subsection{Comparison with State-of-the-arts}

In Table \ref{tab:miniacc}, we compare the performance of our algorithm with several state-of-the-art few-shot learning methods on the \textit{mini}ImageNet and the \textit{tiered}ImageNet. 
On the \textit{tiered}ImageNet and the \textit{mini}ImageNet, our algorithm achieves a more favorable score than other methods on 1-shot classification.
On the \textit{tiered}ImageNet, our solution outperforms other methods on 1-shot and 5-shot classification by a larger margin.
Centroid alignment~\cite{alignment} is slightly better than ours when dealing with 5-way 5-shot tasks on \textit{mini}ImageNet. However, Centroid alignment~\cite{alignment} adopts WRN-28-10 as the backbone, which introduces a lot more parameters than ResNet-12. As is pointed out in ~\cite{closerfewshot}, the parameter size of the backbone has a great impact on the performance of a few-shot learning method. Note that PSN~\cite{projectivesubspace}, DSN-MR~\cite{subspace} and PN+IFSM~\cite{ifsm} are all proposed to optimize the class-level prototypes by learning to form an affine subspace or by an iterative optimization mechanism, respectively. But PSN~\cite{projectivesubspace} and PN+IFSM~\cite{ifsm} cannot be applied to the 1-shot classification task. DSN-MR~\cite{subspace} has more flexibility than PSN~\cite{projectivesubspace} and PN+IFSM~\cite{ifsm}. DSN-MR~\cite{subspace} uses the same backbone as we do, but our method has higher accuracy.


\subsection{Performance in Cross-domain Scenario}

In Table \ref{tab:cross}, we experiment our method on cross-domain tasks, which is trained on \textit{mini}ImageNet and tested on CUB-200-2011. In our experiment, ResNet-12 is used as the backbone, which is the same as the comparable method. Table \ref{tab:cross} shows that our method is the best performer on cross-domain tasks, which means that the episodic prototype generator can benefit in the cross-domain scenario. 
\begin{table}[t]
    \centering
    \caption{\textbf{Few-shot classification accuracy (\%) on a cross-domain task (\textit{mini}ImageNet $\rightarrow $ CUB-200-2011).} $^{\dag}$ indicates that the experimental results are from \protect\cite{closerfewshot}. \textbf{Bold}: the best.}
    \begin{tabular}{r| c c}
    \toprule
        \multirow{1}*{\textbf{Methods}}
         &\multicolumn{1}{c}{5-way 5-shot} \\
    \midrule
    Baseline++$^{\dag}$~\cite{closerfewshot}         & 62.04 $\pm$ 0.76 \\
    MatchingNet$^{\dag}$~\cite{matching}             & 53.07 $\pm$ 0.74 \\

    MAML$^{\dag}$~\cite{maml}                        & 51.34 $\pm$ 0.72 \\
    ProtoNet$^{\dag}$~\cite{protonet}                & 62.02 $\pm$ 0.70 \\
    RelationNet$^{\dag}$~\cite{rn}                   & 57.71 $\pm$ 0.73 \\

    \midrule
    Ours                                    & \textbf{62.60 $\pm$ 0.75} \\
    \bottomrule
    \end{tabular}
    \label{tab:cross}
\end{table}

\subsection{The Effect of Episodic Prototype Generator}

We conduct ablation experiments to verify the effectiveness of our mechanism. All the experiments are performed on 5-way 1-shot and 5-way 5-shot tasks, with \textit{mini}ImageNet being the dataset.

\begin{table}[hb]
    \centering
    \caption{\textbf{Ablation studies on effect of prediction accuracy (\%).}}
    \begin{tabular}{r|c c}
    \toprule
        \multirow{2}*{\textbf{Methods}}
         &\multicolumn{2}{c}{\textbf{5-way}} \\
         \multicolumn{1}{c|}{~}
         &\multicolumn{1}{c}{1-shot}
         &\multicolumn{1}{c}{5-shot} \\
    \midrule
    ProtoNet+Lock       & 59.13 $\pm$ 0.86 & 80.88 $\pm$ 0.58\\
    ProtoNet            & 61.17 $\pm$ 0.82 & 79.86 $\pm$ 0.55 \\
    \midrule
    Ours                &66.68 $\pm$ 0.81 & 82.22 $\pm$ 0.56  \\
    \midrule
    \Targetproto{}  &  89.15 $\pm$ 0.37    & 89.15 $\pm$ 0.37 \\
    
    \bottomrule
    \end{tabular}
    \label{tab:ablation_effectiveness}
\end{table}

In Table \ref{tab:ablation_effectiveness}. We first test ProtoNet+Lock and ProtoNet. The former has frozen backbone parameters, while the latter is learnable during the training phase. 
They both use the arithmetic mean as the \generatedproto{}. Our method is based on ProtoNet+Lock with the feature generator module applied. As can be seen from Table \ref{tab:ablation_effectiveness}, the proposed module has a larger improvement of 7.55\% in 1-shot setting and 1.34\% in 5-shot settings compared with ProtoNet+Lock. 
Besides, the proposed module also shows a larger improvement compared to ProtoNet, which has learnable parameters. This shows the effectiveness of our episodic prototype generator module.

\begin{figure}[t] 
\centering 
\includegraphics[width=0.4\textwidth]{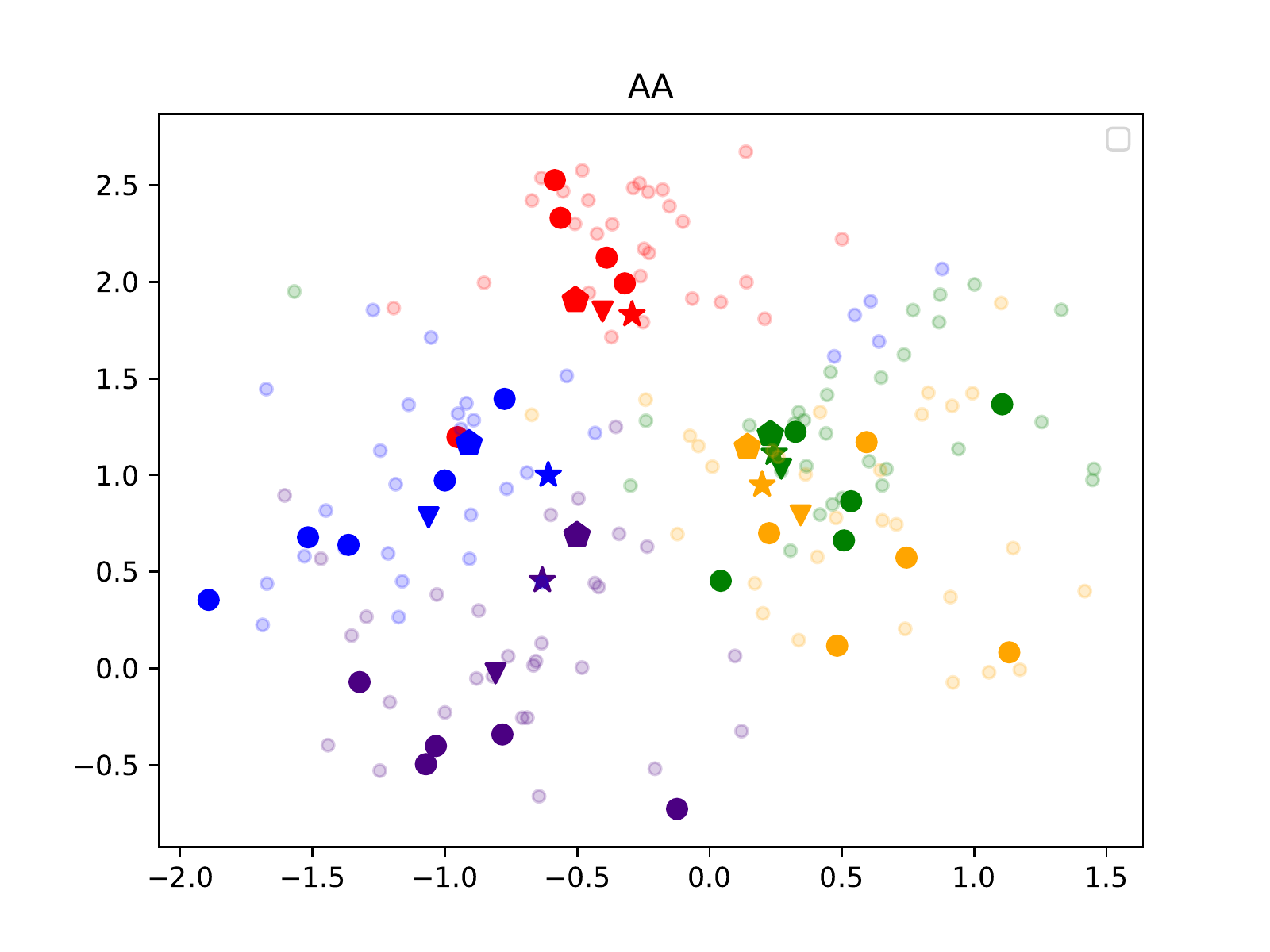} 
\caption{\textbf{Visualization of episodic prototype features and mean prototype features of different models.} Different colors represent categories, star-shaped point represent \targetproto{}, triangles represent mean prototype features of ProtoNet \protect\cite{protonet}, pentagons are \generatedproto{}s generated by our episodic prototype generator module, large dots are features of support samples, and small dots are features of randomly selected samples of the same class.} 
\label{fig_visualization}
\end{figure}

We also report the prediction results using \targetproto{}, which are the ideal results. This experiment uses the \targetproto{}s of each support sample to classify the query samples, and the purpose of reporting this result is to explore the upper bound of the capability of our proposed method. At the same time, we can see that the prediction accuracy can reach 89.15\% when using the \targetproto{} as the \generatedproto{}. There leaves a big gap for improvement, and we believe that the gap can be further reduced by improving the ability of the episodic prototype generator module.

To verify the validity of the proposed method, we visualize the features of the support samples, \targetproto{}s, \generatedproto{}s, and mean prototypes using tsne~\cite{tsne} in Figure \ref{fig_visualization}. The mean prototype feature is generated by ProtoNet~\cite{protonet}, which uses the same backbone as our model, and the \generatedproto{}s are generated by our model.
It shows that there are outliers in the purple and blue classes, resulting in a large gap between the mean prototype of these two classes and their \targetproto{}s, while the \generatedproto{} can be closer to \targetproto{}. It shows that our method can obtain a more robust prototype in the presence of outliers.

The above experiments show that in the presence of outliers, the proposed episodic prototype generator can obtain more robust prototypes and improve the prediction accuracy in the presence of few-shot learning.




\section{Discussion}
In a few-shot learning scenario, the arithmetic mean prototype tends to be influenced by outlier points. Through experimental observations, we find class-level prototype can greatly improve few-shot learning model. In this paper, we propose a simple method to learn class-level prototypes from few support samples. This method is based on the idea that we can generate more reliable prototypes supervised with class-level prototypes. The proposed method can produce competitive results than most metric-based few-shot learning methods, only by improving the prototype. 

\newpage
\bibliographystyle{named}
\bibliography{FSL_Classlevel}
\end{document}


\maketitle

\section{Implementation Detail}
In this section, first, we illustrate the detail of the pre-trained strategy of our network in the training stage and the structure of Transformer in our method.  


\subsection{Pre-trained Strategy} 

In the training stage, we apply additional pre-training strategies to the proposed model. As we can see, all classes in the training set in each dataset (\eg  351 classes in the \tieredimagenet{}) are classified by cross-entropy loss training using Softmax layer attached to the backbone network. Thus we use image augmentations to enhance the generalization ability of the model, such as random cropping, color jittering, and random flipping During this period.  After each epoch, we verify the quality of the pre-trained weights on the model validation set. Specifically, we randomly select 200 1-shot N-way few-shot learning tasks (N = the number of classes in the validation segmentation, for example, 97 classes in the \tieredimagenet{}), supported one instance per class in the set, and 15 query instances per class for evaluation. Based on the second-to-last instance embedding of pre-training weights, the tasks with fewer shots are classified by using the nearest neighbor classifier, and the performance of the backbone was evaluated. The weights with the best classification accuracy for a few pre-trained shots are selected on the verification set, and the embedding trunk is initialized with the pre-trained weights, and then the weights of the whole model are optimized together in the process of model training.


For \miniimagenet{} and \tieredimagenet{} datasets, we set the initial learning rate as 0.1 and the batch size as 128. The schedule table of the optimizer will be adjusted in the validation part of the dataset, in which the schedule table of the optimizer of \miniimagenet{} is [350,400,440,460,480], and that of the \tieredimagenet{} is [400,500,550,580].


\subsection{Transformer}
In this part, we will describe in detail how we use the Transformer architecture \cite{transformer} to implement an episodic prototype generator module $\mathcal{G}_{\gamma}(\cdot)$, in which a self-attention mechanism that takes into account contextual embedding helps with instance embedding adaptation. Specifically, Transformer works by first storing a triple containing query, key, and value. The elements in the query set are the ones that we want to transform. Transformer matches query points to each key by calculating query-key similarities. The proximity of the key to the query point is then used as a weight to weight the corresponding value for each key. The transformed input acts as a residual value that will be added to the input. We introduce the encoding and multi-head encoding in the following. 
\paragraph{Encoding.}According to \cite{transformer}, the encoder consists of two components, one is the self-attention layer, which helps the encoder pay attention to the different dimensional feature of each sample while encoding, and the other is the feed-forward neural network layer, the output of the self-attention layer serves as the input of this layer.

For each different task, query, key, and value vectors will be generated. These three vectors are embedded according to the instance, and they are combined to create three sets of query, key and value, denoted as  $\mathcal{Q}$, $\mathcal{K}$, and $\mathcal{V}$. At the same time, three weight matrices  $\mathcal{W_Q}\in\mathbb{R}^{d_{model}\times d_k}$, $\mathcal{W_K}\in\mathbb{R}^{d_{model}\times d_k}$ and $\mathcal{W_V}\in\mathbb{R}^{d_{model}\times d_v}$ are corresponding to the above three matrices to enhance the flexibility of the encoder.
\begin{align}
    \mathbf{Q} &= [\Phi_{\theta}(x_q); \forall{x_q}\in\mathcal{Q}]^T\mathcal{W_Q} \in\mathbb{R}^{{|\mathcal{Q}|}\times d_k}, \\
    \mathbf{K} &= [\Phi_{\theta}(x_k); \forall{x_k}\in\mathcal{K}]^T\mathcal{W_K} \in\mathbb{R}^{{|\mathcal{K}|}\times d_k},\\
    \mathbf{V} &= [\Phi_{\theta}(x_v); \forall{x_v}\in\mathcal{V}]^T\mathcal{W_V} \in\mathbb{R}^{{|\mathcal{V}|}\times d_v},
\end{align}
where $\Phi_{\theta}(x_*)$ denotes the instance embedding of set *(\eg $\mathcal{Q}$, $\mathcal{K}$, and $\mathcal{V}$), and because there is a one-to-one correspondence between keys and values, that is $|\mathcal{Q}|=|\mathcal{K}|$. The output matrix of the self-attention layer is $\mathbf{Z}$:
\begin{equation}
    \mathbf{Z} = \mathbf{Softmax}(\frac{\mathbf{Q} \mathbf{K}^T}{\mathbf{\sqrt{d_k}}})\mathbf{V},
\end{equation}
where $d_k$ is the dimension of query and key. We need to mention detail in the encoder architecture that there is a residual connection around each component in each encoder.
\begin{align}
    \widetilde{\Gamma}_{x_q} &= \sum_k Z_{:,k},\\
    \Gamma_{x_q} &= \tau(\Phi_{\theta}(x_q)+\widetilde{\Gamma}_{x_q}^T\mathcal{W_{FC}}).
\end{align}

The product of $\mathbf{Q}$ and $\mathbf{K}$ matrices reveals the special proximity between $x_k$ and $x_q$, that is, the attention value, and multiplies it with $\mathbf{V}$ as the weight of the final embedding $x_q$. After that, further conversion $\tau$ is completed by Dropout and Layer Normalization.
\paragraph{Multi-head Encoding.}The expanded encoder is similar to the basic structure, but the sub-layer is further improved self-attention layer by adding a mechanism called \textit{multi-head} attention which can be parallelized, and the performance of the attention layer is improved by extending the ability of the model to focus on different samples and giving multiple \textit{presentation subspaces} of the attention layer. Therefore, under the \textit{multi-head} attention mechanism, we maintain an independent weight matrix $\mathcal{W_Q}$, $\mathcal{W_K}$ and $\mathcal{W_V}$ for each head to generate different 
$\mathbf{Q}$, $\mathbf{K}$ and $\mathbf{V}$ matrices. As above, we multiply $\Phi_{\theta}(x_*)$ by the $\mathcal{W_Q}$, $\mathcal{W_K}$ and $\mathcal{W_V}$ to generate the $\mathbf{Q}$, $\mathbf{K}$ and $\mathbf{V}$. If we use \textbf{H} attention heads, we only need to do \textbf{H} different weight matrix operations to get \textbf{H} different \textbf{Z}. In simple terms, $\mathcal{Q}$, $\mathcal{K}$, and $\mathcal{V}$ are projected through 
H different linear transformations, and finally the different attention results are spliced together
\begin{equation}
    \mathbf{Z^*} = (\underset{i}{\operatorname{Concat~}} \mathbf{Z_i})^T \mathcal{W_O},
\end{equation}
where $\mathbf{Z_i} = \mathbf{Softmax}(\frac{\mathbf{Q}_i \mathbf{K}_i^T}{\mathbf{\sqrt{d_k}}})\mathbf{V}_i$, $i\in[1, H]$, and parameter matrix $\mathcal{W_O}\in\mathbb{R}^{d_v H\times d_{model}}$.

\newpage
\bibliographystyle{named}
\bibliography{reference}